# COHERENT KNOWLEDGE PROCESSING AT MAXIMUM ENTROPY BY SPIRIT

**Wilhelm Rödder, Carl-Heinz Meyer**
**FernUniversität Hagen, Germany**

## Abstract

SPIRIT is an expert system shell for probabilistic knowledge bases. Knowledge acquisition is performed by processing facts and rules on discrete variables in a rich syntax. The shell generates a probability distribution which respects all acquired facts and rules and which maximizes entropy. The user-friendly devices of SPIRIT to define variables, formulate rules and create the knowledge base are revealed in detail. Inductive learning is possible. Medium sized applications show the power of the system.

## 1 INTRODUCTION

Probabilistic reasoning and knowledge representation are a powerful mathematical tool able to reflect vagueness and uncertainty in human thinking. Within the last decade a growing number of scientists have agreed that a joint distribution $P$ on a set $V$ of discrete variables with finite domains is a good means to represent complex dependency and independence relations amongst them. So consequently it is a perfect knowledge base cf. [8], [11]. In this environment reasoning is understood as the calculation of conditional probabilities for event-conclusions given event-premises.

Since for real world applications the number of elementary events in the probability space counts in billions a significant effort had to be made for an efficient storage of the distribution and a fast calculation of all probabilities.

Conditional independence assumptions allow a decomposition of $P$ and break it down to a considerably reduced number of (un-) conditioned probabilities. Nevertheless for a high dimensional $P$ the dependency and independence relations must be well organized to be consistent and yet determine a unique distribution. Graphs on $V$ formalize the „skeleton" which together with (un-) conditioned probabilities fixes the global $P$.

Bayes-Networks, special Directed Acyclic Graphs (DAGs), provide the most accepted and already standardized skeleton. Undirected and mixed graphs allow a richer structure than DAGs, but at the moment are not yet standard [14].

To make these factorized distributions $P$ admissible to local computations, they must be enriched by further conditional probabilities; the resulting graphical structure is a hypertree [6].

Yet for high dimensional distributions the construction of a Bayes-Net is not trivial at all and requires thousands of (conditional) probabilities. Often high-handed independence assumptions are necessary to form a manageable graph. These lacks might deter inclined users from the application of such a probabilistic knowledge based system.

SPIRIT is an Expert System Shell for building and applying knowledge bases on sets of discrete variables which differs significantly from conventional systems:

- Knowledge acquisition in SPIRIT is performed by processing facts and rules. Facts and rules are (un-) conditioned propositional sentences with a rich syntax and their respective desired probabilities.
- No explicit independence structure must be specified, avoiding the troublesome construction of a Bayes-Net or similar frames.
- Any query about possible conclusions from hypothetical premises allows an equally rich syntax as for knowledge acquisition.

The way SPIRIT acquires and represents the expert-knowledge was first proposed 1983 by Cheeseman in [1]. A similar approach to this form of knowledge representation can be found in the book of Hajek et al. (cf. [3], sec. 5). Some background in Probabilistic Logic (cf. [9]) would also be helpful for a better understanding of this paper. It's subject is the presentation of the shell SPIRIT and its way of working with uncertain knowledge. So only a brief introduction to the theory of coherent knowledge processing is given in chapter 2. The shell's features are detailed in chapter 3. We pass the steps Knowledge Acquisition (3.1), Knowledge Processing (3.2), Knowledge Representation (3.3), Queries and Response (3.4) in succession. All items are enriched by examples.

Finally we report on some medium sized applications of SPIRIT in chapter 4. Examples.

## 2 KNOWLEDGE PROCESSING IN SPIRIT

To describe briefly the theoretical background of SPIRIT we need the following mathematical prerequisites. The knowledge base consists of a finite set of variables $V=\{V_1,...,V_n\}$ with finite domains and a probability distribution $P$ on the field of all events on $V$. Events are identified with propositional sentences on literals $V_i=v_i$ (where $v_i$ is a realization of $V_i$) built by negation, conjunction and disjunction. If $S$ is such a sentence, its probability is $p(S)=\sum_{v\subset S} p(v)$, where $v$ runs through all complete conjunctions $v$ in the canonical disjunctive normal form of $S$.

To generate a distribution $P$ for representing knowledge, we first assign desired probabilities $x_f$ or $x_r$ to facts or rules. Here a fact $F$ is a propositional sentence as defined above and a rule is an expression $F_2|F_1$ with $F_1,F_2$ being facts. The assignments are understood as an imperative: find a $P$ for which $P(F)=x_f$ or $P(F_2|F_1)=x_r$; they are considered to be (un-) conditioned probabilities.

Consequently, the first step of knowledge acquisition in SPIRIT consists of the formulation of a set of such facts and rules as well as the assignment of consistent desired probabilities:

$$p(F_{2i}|F_{1i}) = x_i, \ i=1...m \qquad (1)$$

(Since $F_{1i}$ might be tautological, (1) includes both facts and rules).

Of course, in general (1) does not determine a unique $P$ but rather a set of distributions. This is so because the number of rules in (1) is rather small and in most cases, the set of rules will be insufficient to define a skeleton like a Bayes-Net. The concept of entropy and relative entropy will help to remove this lack.

For two distributions $P$ and $P_0$ on $V$ the relative entropy of $P$ with respect to $P_0$ is defined as

$$R(P,P_0) = \sum_v p(v) \, ld\left(\frac{p(v)}{p_0(v)}\right), \qquad (2)$$

with $ld$ denoting the binary logarithm.

If $P_0$ is the uniform distribution, $R(P,P_0)$ becomes equal to

$$H(P) = -\sum_v p(v) \, ld \, p(v) \qquad (3)$$

and is called (absolute) entropy of $P$.

A distribution $P=P^*$ which minimizes (2) subject to (1) has desirable properties, such as that

- it preserves the dependency structure of $P_0$ as far as possible when forced to satisfy (1), see [10].
- it minimizes additional information in $P^*$ beyond that already existing in $P_0$, cf. [3],[12].

We take the convincing arguments in the referred papers as a good reason to solve the following problem when acquiring knowledge additional to $P_0$ by facts and rules:

$$Min \sum_v p(v) \, ld\left(\frac{p(v)}{p_0(v)}\right)$$

$$\text{s.t.} \qquad p(F_{2i}|F_{1i}) = x_i, \ i=1...m \qquad (4)$$

In SPIRIT we solve (4) iteratively applying only one restriction at a time, thus generating a sequence of distributions $P_0, P_1,...,P_{k-1}, P_{k,...}$ for which each $P_k$ has minimal relative entropy with respect to $P_{k-1}$. It is well known that this procedure converges towards $P^*$ [2]. The application of a single restriction on $P_{k-1}$ to get $P_k$ is an immediate consequence of the Kuhn-Tucker theorem. More precisely: if $P_{k-1}$ is the actual distribution on $V$ and if the next rule to be applied is $p(F_{2i}|F_{1i})=x_i$, $(i = k \ \text{mod} \ m)$, then the Kuhn-Tucker conditions yield:

$$p^k(v) = \begin{cases} p^{k-1}(v)\dfrac{1-a^*}{p^{k-1}(\overline{F_1})}, & \forall v \subset \overline{F_1} \\[2ex] p^{k-1}(v)\dfrac{(1-x)a^*}{p^{k-1}(F_1\overline{F_2})}, & \forall v \subset F_1\overline{F_2} \\[2ex] p^{k-1}(v)\dfrac{xa^*}{p^{k-1}(F_1F_2)}, & \forall v \subset F_1F_2 \end{cases}$$

Here barring indicates negation and we write $F_1F_2$ for $F_1 \wedge F_2$. $a^*$ obviously is the posterior probability of the premise $F_1$ and may be calculated according to:

$$a^* = \frac{p^{k-1}(F_1F_2)^x \, p^{k-1}(F_1\overline{F_2})^{1-x}}{p^{k-1}(F_1F_2)^x p^{k-1}(F_1\overline{F_2})^{1-x} + p^{k-1}(\overline{F_1})x^x(1-x)^{1-x}}$$

Up to now we developed how entropy substitutes in an ideal way the Bayes-Net as skeleton. It remains to show how local computability is guaranteed in SPIRIT.

Each rule (or fact) involves a certain group of variables and generally implies dependency relations among them. So each such set of variables forms in a natural way a cluster for which the marginal distribution should be fully stored. The set of all clusters constitutes a hypergraph on $V$, in general a cyclic one. SPIRIT therefore seeks for an acyclic hypergraph – a hypertree – in such a way that each of the above clusters as a whole is contained in a hyperedge. The corresponding algorithms are similar to those for triangulating a graph. They all start with the – in general cyclic – hypergraph, whose hyperedges are the clusters of variables mentioned above. Then they add further variables to clusters as to "fill in" undesired cycles in the hypergraph. The idea behind clustering might be a maximum cardinality search or a minimum fill-in concept, cf. [5], [13]. We realized an algorithm of either type. As both methods are heuristics, the efficiency of the used method depends on the original structure.

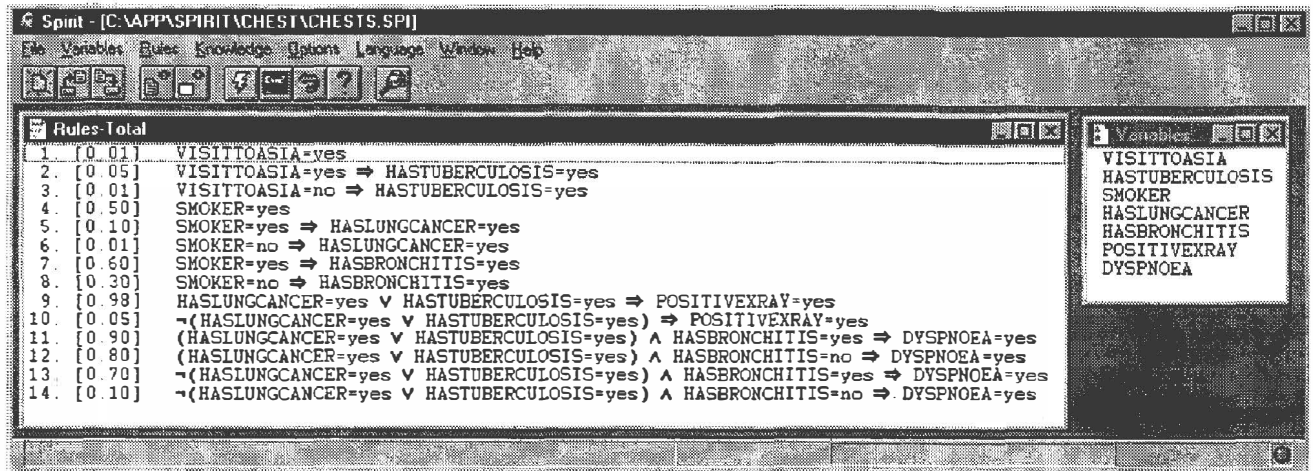

**Figure 1:** Facts and rules of the chest clinic (the meaning of an expression like „[0.10] $F_1$=>$F_2$" is: $P(F_2|F_1)=0.1$)

For the resulting hyperedges the marginal distributions are fully stored. The following synonymous names are established for these hyperedges: belief universes, Local Event Groups (LEGs), etc. So global computations are reduced to local computations: modifications in a hyperedge can be „propagated" throughout the hypertree, cf. [6]. The compact presentation of mathematical prerequisites was necessary to understand the different devices to be shown in the next chapter.

## 3 THE SHELL AND ITS FEATURES IN DETAIL

### 3.1 KNOWLEDGE ACQUISITION

One of the best ways to explain the process of knowledge acquisition is perhaps „learning by example". The following example is adapted from a well known Bayes-Network (for a detailed description cf. [6]) and represents some fictitious medical knowledge. We cite:

**Example 1:**

> „Shortness of Breath (dyspnoea) may be due to tuberculosis, lung cancer or bronchitis, or none of them, or more than one of them. A recent visit to Asia increases the chances of tuberculosis while smoking is known to be a risk factor for both lung cancer and bronchitis. The results of a single chest X-ray do not discriminate between lung cancer and tuberculosis, as neither does the presence or absence of dyspnoea."

After declaring the types and the values of 7 variables (shown on the right side in Figure 1) it is possible to specify some probabilistic facts and rules (shown on the left side in Figure 1)

The facts are propositional sentences and the rules link two such sentences by „⇒". Facts and rules have to meet certain syntax demands, see Figure 2.

The propagation of a rule can follow either the philosophy 'float' or 'ground'. A floating rule in SPIRIT is handled as in (4), see chapter 2. In this version the user does not impose any restriction upon the probability of the premise – it might float. If he grounds it, he wants the probability of the premise to remain unchanged.

**Example 2:** The probability distribution $P^*$ which represents the rule $P(B|A)=1$ for boolean variables $A$, $B$ is shown for 'float' and 'ground' in the following table (e.g. $ff$ stands for $A = f \wedge B = f$).

| $ff$ | $ft$ | $tf$ | $tt$ | | |
|------|------|------|------|---|---|
| 0.25 | 0.25 | 0.25 | 0.25 | uniform | $P_0$ |
| 0..333.. | 0..333.. | 0.00 | 0..333.. | 'float' | $P^*$ |
| 0.25 | 0.25 | 0.00 | 0.5 | 'ground' | $P^*$ |

```
<Rule> ::=<Fact>⇒<Fact>
<Fact> ::={¬}<Atom>|(<Fact>){∧|∨<Atom>|(<Fact>)}
<Atom> ::=<Variable>(=|≠|<¹|>¹|∈|∉<Value>|<Valuelist>²}    1:Ordinal 2:Non-Boolean
<Valuelist> ::=<Value>{,<Value>}
```

**Figure 2:** Syntax of a probabilistic facts and rules

Observe that both philosophies respect the desired rule probability, but only 'ground' maintains $P^*(A=t)=P_0(A=t)=0.5$. In SPIRIT the more natural application of rules is that of type 'float', but if the user wants to preserve the condition structure in the premise he should use the option 'ground'.

**Example 3:** The probability distribution $P^*$ which represents the rule $P(C|A \wedge B)=0.9$ is shown for 'float' and 'ground' in the following table (e.g. *fff* stands for $A = f \wedge B = f \wedge C = f$).

| *fff* | *fft* | *ftf* | *ftt* | *tff* | *tft* | *ttf* | *ttt* | |
|--------|--------|--------|--------|--------|--------|--------|--------|----------|
| 0.1250 | 0.1250 | 0.1250 | 0.1250 | 0.1250 | 0.1250 | 0.1250 | 0.1250 | 'uniform' |
| 0.1354 | 0.1354 | 0.1354 | 0.1354 | 0.1354 | 0.1354 | 0.0187 | 0.1687 | 'float' |
| 0.1250 | 0.1250 | 0.1250 | 0.1250 | 0.1250 | 0.1250 | 0.0250 | 0.2250 | 'ground' |

Note that both philosophies respect the desired rule probability, yet 'float' introduces a slight dependency between $A$ and $B$ (e.g. $P(B|A) = 0.4090 \neq 0.5$), whereas 'ground' does not.

A 'float' rule is more natural in that it puts less information into $P^*$ than a 'ground' rule. It makes its duty with the highest entropy and so is coherent with (and only with) the facts and rules formulated by the user. Additional independence assumptions should be incorporated in the distribution only if explicitly wanted – as it means extra information. The entropy for the 'ground' distribution in the above example 3 is lower than that of 'float'. In this sense a 'ground' rule might cause incoherent independence structures. We implemented the option 'ground' only to facilitate the creation and import of Bayesian - Networks.

Once all variables and rules are specified, the shell is ready for the computation of the knowledge base.

### 3.2  KNOWLEDGE PROCESSING

The proper knowledge processing consists of the construction of the hypertree, mentioned in chapter 2, as well as the iterative calculation of the joint distribution.

To construct the hypertree, SPIRIT offers optional algorithms [5], [13]. Their efficiency might vary for different knowledge bases. Once the hypertree is determined the shell starts an iterative process of applying all facts and rules to the actual distribution. The user is informed about the number of full iterations (all rules) and the actual rule applied. Further information concerns entropy: the relative entropy between the distributions after and before the rule application (c.f. chapter 2), the absolute entropy of the uniform distribution minus the sum of all processed relative entropies, and the actual absolute entropy of the joint distribution.

SPIRIT detects whether

- the knowledge base is ready or
- inconsistent rules have been supplied by the user. In this case the shell supports the revision of inconsistencies.

'Alpha-Learning' allows inductive learning from an arbitrarily-sized sample of all variables' realizations by the knowledge base. In each hyperedge every cell in the contingency-table is actualized by the formula $p_{new}=(1-\alpha)p_{old}+\alpha f$, where $p_{old}$ is the probability before alpha-learning and $f$ the frequency in the sample. $0 \le \alpha \le 1$ is a weight which must be suitably chosen by the user. The knowledge base then applies $p_{new}$ to a set of previously defined rules and goes through the learning process again. The idea behind Alpha-learning is to adjust subjective by „objective" probabilities.

The reverse process is also possible, namely to draw a sample which follows the joint distribution.

Once the knowledge base is completed -either by rules and their specified probabilities or by rules 'filled' with frequencies from a sample- it is ready for queries and responses. Before explaining that feature we will show how the shell SPIRIT informs the user about the acquired knowledge.

### 3.3  KNOWLEDGE REPRESENTATION

SPIRIT has two main features to represent the acquired information: the Structure Graph and the Dependency Graph.

The structure graph visualizes the hypertree, generated by algorithms similar to those used for triangulation. The vertices of the graph are the hyperedges of the hypertree, the edges of the graph have a somewhat complicated interpretation. The interested reader might consult [4]. The structure graph informs about the quality of the triangulation.

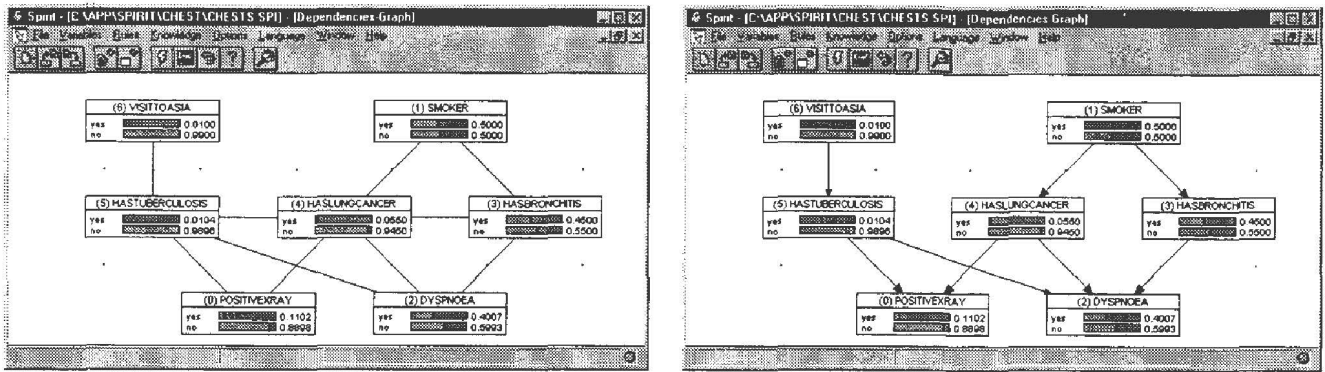

**Figure 3:** Dependence and mixed graph of example 1

The Dependency-Graph is an undirected graph constructed as follows:

- for each variable generate a vertex
- two vertices are linked by an edge, if the variables appear in the same rule.

The graph has the Local Markov Property [11]. Bar-charts displayed in the vertices show the actual marginal distribution of the variable. It is possible to switch to a mixed graph constructed as follows:

- for each variable generate a vertex
- link $v_1$ and $v_2$ by an arrow, if $v_1$ is in the premise and $v_2$ in the conclusion of the same rule
- link vertices $v_1, v_2$ by an edge if they appear both in the conclusion of the same rule.

For the example developed earlier Figure 3 shows the dependence graph on the left side and the mixed form on the right side. Confirm that the rules in Figure 1 do not generate undirected edges.

In this chapter we revealed properties of the dependence graph as it serves for visualization of the knowledge base. The graph might be used for consultations, too. This aspect and complex queries in SPIRIT are developed in the next chapter.

### 3.4    QUERIES AND RESPONSES

SPIRIT provides two forms of consulting: simple questions and complex queries. Simple questions are available in any conventional probabilistic knowledge base, complex queries are a sophisticated option in SPIRIT which will be developed below.

Simple questions are performed in the Dependency-Graph by instantiation of one or more variables' values. This temporary modification is propagated throughout the whole knowledge base.

Since this kind of query is common to any probabilistic knowledge base we omit an example and further discussion.

Complex queries are frequently necessary when consulting a (probabilistic) knowledge base. In SPIRIT a set of hypothetical temporary facts and/or rules can be formulated and then — given these circumstances — further „imperative" facts or rules might be evaluated. Both, hypothetical and imperative facts and rules allow the same rich syntax as during knowledge acquisition. We continue the example 1 of chapter 3.1:

**Example 1 (continued):** Let us assume that there is a strong evidence (p=0.9) for a patient to have bronchitis or lung cancer. We want to know the probability that he or she is a smoker. To answer this query, SPIRIT calculates the conditional probability given virtual evidence and finds the result in Figure 4. What happens in a query mathematically, is to solve instead of (4) the new problem:

$$\text{Min} \sum_v p(v) = ld\left(\frac{p(v)}{p^*(v)}\right) \qquad (5)$$

s.t.

hypothetical facts and rules are fulfilled.

After solving problem (5), with solution $P^{**}$, we evaluate the expressions on the right side (Conclusio) shown in Figure 4 under $P^{**}$.

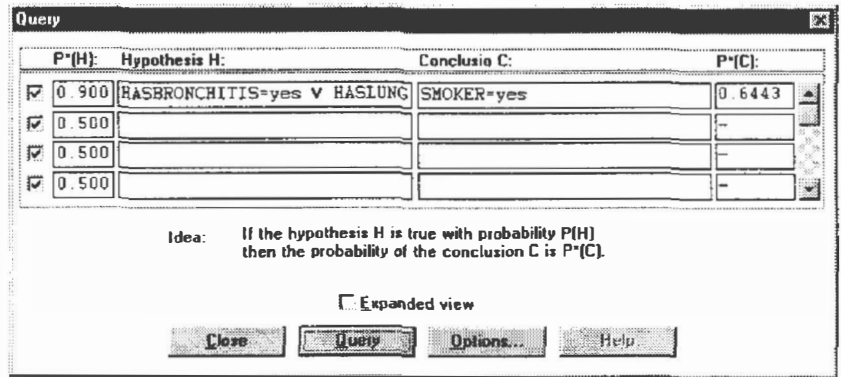

**Figure 4:** A typical complex query in SPIRIT

### 3.5    FACTS AND FEATURES OF SPIRIT

The system SPIRIT consists of two parts: the kernel and the Graphical User Interface (GUI). The kernel is programmed in Standard-C++ with an interface for application-programmers. It should run under any operating system with a suitable C-Compiler. The GUI is developed for WINDOWS 95™ and supports at the moment three languages: German, English and Portuguese. There are also Import-Filters available for some of the common Belief-Network-Description Languages, e.g. Microsoft™ DSC-Format and HUGIN™ NET-Format. A restricted Demo-Version may be found on WWW under the following URL:

<http://www.fernuni-hagen.de/BWLOR/sp_demo.htm>.

### 4.    CONCLUSION

Many applications of SPIRIT are on the way:

A medium sized example is BAGANG which allows medical diagnoses by Chinese medicine. It actually involves 200 rules and 50 variables.

SPIRIT is used to train efficiently personnel in a Brazilian management simulation game and, a small-sized knowledge base supports decisions of a German bank concerning credit-worthiness.

All Bayes-Net-type structured data-sets available by Internet where tested in SPIRIT. In the case of the very famous probabilistic model „blue" baby [7] the rich syntax of SPIRIT allowed a considerable reduction from the original 341 conditioned and unconditioned probabilities to 164 facts and rules. Our unprofessional tests (minimum medical knowledge) resulted in very similar diagnoses as for the original system. Concerning the performance of SPIRIT, the calculation of the joint distribution as in (4) takes the most time. So it took approximate 50 sec. to learn the facts and rules of the 'blue-baby' on a Standard-PC. The response time of simple questions is equal to that of 'conventional' systems, for complex queries it counts in seconds.

### References

[1] P. Cheeseman (1983). *A method of computing generalized Bayesian probability values for expert systems.* Proc., 6.th Intl. Joint Conf. on AI (IJCAI-83), Karlsruhe, Germany, 198-202.

[2] I. Csiszár (1975). *I-divergence Geometry of Probability Distributions and Minimization Problems,* Ann. Prob. 3, 146-158.

[3] P. Hájek, T. Havránek; R. Jiroušek (1992). *Uncertain Information Processing in Expert Systems,* CRC-Press, Boca Raton, Florida.

[4] V. Jensen, F. Jensen (1994). *Optimal junction trees, in: Proceedings of the Tenth conference on Uncertainty in Artificial Intelligence.* Morgan Kaufmann Inc., 360-366.

[5] U. Kjaerulff (1990). *Triangulation of graphs — algorithms giving small total state space.* Technical Report R90-09, Dept. of Mathematics and Computer Science. Aalborg University.

[6] S. L. Lauritzen, D. J. Spiegelhalter (1988). Local *Computations with probabilities on graphical structures and their application to expert systems.* J. Roy. Stat. Soc. Ser. B, 50, 157.

[7] S. L. Lauritzen, B. Thiesson, D. J. Spiegelhalter (1994). *Diagnostic systems by model selection: a case study,* Lecture Notes in Statistics, 89. Springer, 143-152.

[8] R. E. Neapolitan (1990). *Probabilistic Reasoning in Expert Systems — Theory and Algorithms.* Jon Wiley & Sons, New York.

[9] N.J. Nilsson (1986). *Probabilistic Logic.* Artificial Intelligence 28 (no.1): 71 - 87.

[10] J. B. Paris, A. Vencovská (1990). *A Note on the Inevitability of Maximum Entropy*, Int. J. Approximate Reasoning, 4, 183-223.

[11] J. Pearl, (1988). *Probabilistic Reasoning in Intelligent Systems*. Morgan Kaufmann, San Mateo, California.

[12] J. E. Shore, R. W. Johnson (1980). *Axiomatic Derivation of the Principle of Maximum Entropy and the Principle of Minimum Cross Entropy*. IEEE Trans. Inform. Theory
IT - 26,1,pp 26-37.

[13] R. E. Tarjan, M. Yannakakis (1984). *Simple linear-time algorithms to test chordality of graphs, test acyclicity of hypergraphs and selectively reduce acyclic hypergraphs*, SIAM J. Comp.,13, 566-579.

[14] J. Whittaker (1990). *Graphical Models in Applied Multivariate Statistics*, John Wiley & Sons, New York.